\pdfoutput=1
\documentclass[11pt]{article}

\usepackage{emnlp2021}

\usepackage{times}
\usepackage{latexsym}
\usepackage[T1]{fontenc}
\usepackage[utf8]{inputenc}

\usepackage{microtype}
\usepackage{graphicx}
\usepackage{times}
\usepackage{url}
\usepackage{subfigure}
\usepackage{longtable}
\usepackage{booktabs}

\usepackage{amsmath}
\usepackage{amssymb}

\newcommand*\samethanks[1][\value{footnote}]{\footnotemark[#1]}

\title{Bilingual Alignment Pre-Training for Zero-Shot Cross-Lingual Transfer}

\author{Ziqing Yang$^1$\thanks{\ \ Equal contribution.}, Wentao Ma$^1$\samethanks, Yiming Cui$^{2,1}$,  Jiani Ye$^1$, 
\textbf {Wanxiang Che$^2$, Shijin Wang$^{3,4}$}\\
{$^1$Joint Laboratory of HIT and iFLYTEK (HFL), iFLYTEK Research, China}\\
{$^2$ Research Center for SCIR,} {Harbin Institute of Technology, Harbin, China}\\
{$^3$iFLYTEK AI Research (Hebei), Langfang, China} \\
{$^4$State Key Laboratory of Cognitive Intelligence, iFLYTEK Research, China}\\
{\tt$^{1,3,4}$\{zqyang5,wtma,ymcui,jnye,sjwang3\}@iflytek.com}\\
{\tt $^2$\{ymcui,car\}@ir.hit.edu.cn}\\
}

\begin{document}
\maketitle
\begin{abstract}
Multilingual pre-trained models have achieved remarkable performance on cross-lingual transfer learning. Some multilingual models such as mBERT, have been pre-trained on unlabeled corpora, therefore the embeddings of different languages in the models may not be aligned very well. In this paper, we aim to improve the zero-shot cross-lingual transfer performance by proposing a pre-training task named Word-Exchange Aligning Model (WEAM), which uses the statistical alignment information as the prior knowledge to guide cross-lingual word prediction. We evaluate our model on multilingual machine reading comprehension task MLQA and natural language interface task XNLI. The results show that WEAM can significantly improve the zero-shot performance.
\end{abstract}

\begin {figure*} [ht!]
  \centering \small
  \includegraphics [width= 0.8\textwidth] {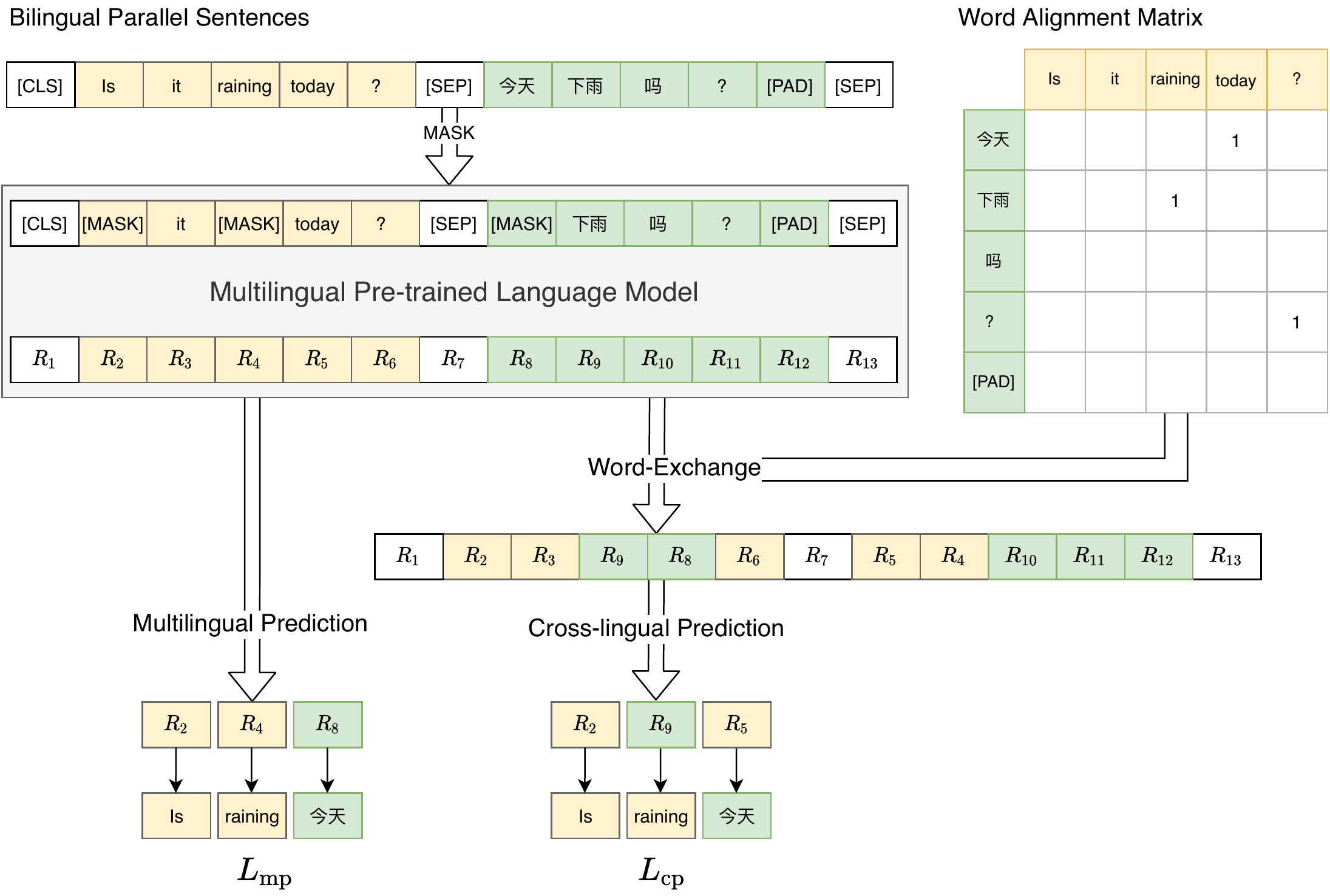}
  \caption{\label{alm-example} \textbf{}  An overview of the Word-Exchange Aligning Model (WEAM). For each language pair, There are two tasks. The multilingual prediction task predicts the masked tokens. The cross-lingual prediction task utilizes a word alignment matrix to swap the representations of aligned words in parallel sentences, then predicts the masked tokens in the swapped sentences.}
\end {figure*}

\section{Introduction}
Large-scale multilingual pre-trained language models such as mBERT \citep {devlin-2019-bert}, XLM \citep{conneau-2019-cross} and XLM-R \citep{conneau-2020-unsupervised} have shown significant effectiveness in transfer learning on various cross-lingual tasks. 
The pre-training methods of the multilingual language models can be divided into two groups: unsupervised pre-training like Multilingual Masked Language Model (MMLM) \cite {devlin-2019-bert, conneau-2020-unsupervised}, and supervised pre-training like Translation Language Model (TLM) \cite{conneau-2019-cross}. In the MMLM, the model predicts the masked tokens with the monolingual context; in the TLM, the model can attend to both the contexts in the source language and target language. Variations of TLM model can be found in \citet{huang-2019-unicoder, chi-etal-2021-infoxlm, DBLP:journals/corr/abs-2012-15674}.

While it is possible for the model to learn the alignment knowledge by itself, some works have investigated injecting prior knowledge to the model to help it to align better.
\citet {cao-2020-multilingual} proposed a bilingual pre-training model for mBERT, where it identifies matched word pairs in parallel bilingual corpora using unsupervised standard techniques such as FastAlign \citep {dyer-2013-simple}, and aligns the contextual representations between the matched words with a similarity loss function. 

The previous works focus on aligning the contextual representations of the pre-trained models.
In this paper, we propose a new cross-lingual pre-trained model called Word-Exchange Aligning Model (WEAM). Different from previous works, we align the static embeddings and the contextual representations of different languages in the multilingual pre-trained models.

Specifically, in the pre-training stage, we first use FastAlign to identify bilingual word pairs in parallel bilingual sentence pairs as our prior knowledge. Then we randomly mask some tokens in the bilingual sentence pairs.
For each masked token, WEAM performs two kinds of predictions: a multilingual prediction and a cross-lingual prediction.
The multilingual prediction task predicts the original masked word in the standard way. while the cross-lingual task predicts the corresponding word from the representations in the other language.
For example, if the words \textit{apple} and \textit{Apfel} (German for \textit{apple}) appear in the the English--German parallel sentence and \textit{apple} is masked in the sentence, WEAM takes the representation of the masked \textit{apple} and \textit{Apfel} for multilingual prediction and cross-lingual prediction respectively to recover the original word \textit{apple}.

Through the two ways of prediction, both the contextual representations from the last transformer layer and the static embeddings from the embedding layer can be aligned. We evaluated our method on the word-level machine reading comprehension task MLQA \citep {lewis-2019-mlqa}
and the sentence-level classification task XNLI \citep {conneau-2018-xnli}.
The results show that WEAM significantly improves the cross-lingual transfer performance.

\section{Methodology}

 \begin{table*}[t]
        \begin{center} \small
        \begin{tabular}{l cccccc}
        \toprule
        Model  & en & es & de & zh & AVG(all) & AVG(zero-shot)  \\ 
         \midrule
         \emph{Translate-Train}\\
         mBERT \dag  & 65.2/77.7 & 37.4/53.9 & 47.5/62.0 & 39.5/61.4 & 47.4/63.8 & 43.0/60.3 \\
         mBERT (ours) & 67.3/80.3 & 48.4/67.1 & 49.1/63.5 & 42.8/63.6 & 51.9/68.6 & 48.1/65.7 \\
         \midrule
        \emph{Zero-Shot} \\
        mBERT \dag & 65.2/77.7 & 46.6/64.3 & 44.3/57.9 & 37.3/57.5 & 48.4/64.4 & 44.2/61.0 \\
        mBERT+TLM & \bf 66.8/80.0 & 47.7/65.7  & 48.4/63.1  & 40.1/62.0 & 50.7/67.7 & 46.7/64.6  \\
        mBERT+WEAM  & 66.7/79.7 &\bf  49.6/67.8 &\bf  49.7/64.3 & \bf 41.7/63.7 & \bf 51.7/68.9 & \bf 48.2/66.2 \\
        \bottomrule
        \end{tabular}
        \end{center}
        \caption{\label{result-mlqa} EM/F1 scores on the test set of MLQA dataset. The results with $\dag$ are taken from  \citet {lewis-2019-mlqa}. \textit{AVG(all)} is the average scores on all languages. \textit{AVG(zero-shot)} is the average scores on the languages excluding English.}
        \end{table*}

\subsection {Translation Language Model}
We first briefly describe the Translation Language Model (TLM) \cite{conneau-2019-cross}.
Like MMLM in  \cite{devlin-2019-bert}, TLM performs the masked word prediction task, where it randomly masks some words and predicts the original ones within a parallel sentence pair.
For each masked word, the model can either attend to the surrounding words or the translated context in the other language, encouraging the model to align the words in different languages.

\subsection {Word-Exchange Aligning Model}
\label {sec: method}
Our proposed method WEAM is based on the multilingual pre-trained model and consists of two tasks: the multilingual prediction task and the cross-lingual prediction task, as shown in Figure \ref{alm-example}.

{\noindent{\bf Multilingual Prediction.}\label{mp} In the multilingual prediction, we randomly mask tokens in the bilingual parallel sentences and predict the original tokens with the outputs from the last transformer layer. unlike TLM,  we did not reset the position embeddings or add the language embeddings, so the distinction between languages will be purely learned from the token embeddings.
We construct the inputs and obtain the representations for a source-target sentence pair $ \left\langle{S, T} \right\rangle$ as
\begin {gather}
X = \texttt{[CLS]}  S \texttt{[SEP]} T  \texttt{[SEP]}  \\
H =  \texttt{Encoder}(X)
\end {gather}
where $X$ is the token sequence and $H \in \mathbb{R}^ {m\times h}$  is the output from the last transformer layer of the pre-trained model \texttt{Encoder}; $m$ is the max sequence length and $h$ is the hidden size.
For a masked token ${X_i}$, we predict the original token $w_i$ with the corresponding representation
\begin {align}
 &H'_i= \delta( W_1 \cdot H_i + b_1)\\
 &p(X_i=w_i| {H}'_i) = \nonumber \\
 &\quad \frac {\exp(\texttt{linear}(H'_i) \cdot e_i) }{ \sum_{k=1}^{\lvert\mathcal{V}\rvert} \exp(\texttt{linear}(H'_i) \cdot e_k)}
\end {align}
where $\delta$ is the GELU activation \cite{hendrycks-2016-gaussian},  $\texttt{linear}(\cdot)$ is a linear layer, $H_i$ is the token representation for  ${X_i}$, as given by Eq (2). $\lvert\mathcal{V}\rvert$ is the vocabulary size. $e_i$ is the emebdding vector of token $w_i$. 

{\noindent{\bf Cross-lingual prediction.} In the cross-lingual prediction, we predict the masked tokens with the representations from the other language. Specifically, we first use FastAlign to construct an alignment words set from parallel sentences $ \left\langle{S, T}  \right\rangle$. We denote the words set as $d(s,t) = \{(i_1, j_1),...,(i_n, j_n)\}$, where $i$ is the word index of source language in the input sequence, $j$ is the word index of the target language. $n$ is the number of word pairs in the sentence pair. Then we generate effectively code-mixed representations by exchanging the positions of each word pair in parallel sentences. We denote the exchange operation with an off-diagonal matrix $A\in\{0,1\}^{m\times m}$:
\[
A(i, j) = 
	\begin{cases}
	1, &\text{if \{$(i,j) ~ or ~ (j, i) \} \in d$}\\
	0, &\text {otherwise}
	\end{cases}
\]
We take $A$ as the transformation matrix to construct the word-exchange representations ${H}^{'}$, which is calculated by
\begin {gather}
H' = A^ \mathrm{ T }  \cdot H  \\
\tilde{H} = W_2 \cdot H' + b_2
\end {gather}
We have applied another linear transformation on ${H}^{'}$ and obtained $\tilde{H}$.
Lastly, we conduct the masked word predictions on $\tilde{H}$ similar to the multilingual prediction: 
\begin {align}
&\tilde{H}'_i = \delta( W_3 \cdot \tilde{H}_i + b_3)\\
& \tilde{p}(X_i = w_i| \tilde{H}'_i) =\nonumber\\
&\quad \frac {\exp(\texttt{linear}(\tilde{H}' _i) \cdot e_i) }{ \sum_{k=1}^{\lvert\mathcal{V}\rvert} \exp(\texttt{linear}(\tilde{H}'_i) \cdot e_k)}
\end {align}
If the word $w_i$ is paired with word $w_j$,
what the cross-lingual prediction does is predicting $w_i$ with the contextual representation of $w_j$. In this way we are align the embedding of $w_i$ ($e_i$) with the contextual representation of $w_j$ ($H_j$).

{\noindent{\bf Pre-training Objective.} 
Given a bilingual parallel corpus $\mathcal{D}$,  we train the multilingual model with the cross-entropy loss. Based on the discussion above, the objective function of pre-training consists of multilingual part $L_{\text{mp}}$ and cross-lingual prediction part $L_\text{{cp}}$.
Let $\Theta$ denote the parameters of the model, then the objective function $L(\mathcal{D}, \Theta)$ can be formulated as
\begin {gather}
L_{\text{mp}} = -\sum\nolimits_{i=1}^M \log (p(w_i)) \\
L_{\text{cp}} = -\sum\nolimits_{i=1}^M \log (\tilde{p}(w_i)) \\
L(\mathcal{D}, \Theta ) = L_{\text{mp}} +  \lambda L_{\text{cp}}
\end {gather}
where $M$ is the number of masked tokens in the instance, $p(w_i)$ and $\tilde{p}(w_i)$, given by Eq.(6) and Eq.(8), are the predicted probability of the masked token $w_i$ over the vocabulary size, $\lambda$ is a hyper-parameter to balance $L_{\text{mp}}$ and $L_{\text{cp}}$.

  \begin{table*}[t]
        \begin{center}\small
        \begin{tabular}{lcccccc}
          \toprule
        Model  & en & es & de & zh & AVG(all) & AVG(zero-shot)  \\ 
         \midrule
        \emph{Translate-Train} \\
         mBERT & 82.1 & 77.8 & 75.9 & 75.7  & 77.9 & 76.5 \\
         \midrule
         \emph{Zero-Shot} \\
        mBERT\dag & 82.1 & 74.3 & 71.1 & 69.3 & 74.2 & 71.6 \\
        Word-aligned BERT\dag   & 80.1 & 75.5 & 73.1 &  -  &  - & - \\
        mBERT+TLM &  82.0 & 75.0   & 73.5  & 73.1  &  75.9  & 73.9 \\
        mBERT+WEAM  &\bf 82.6 & \bf 76.4  &\bf  74.5 & \bf 74.4  & \bf 77.0 & \bf 75.1\\

        \bottomrule
        \end{tabular}
        \end{center}
        \caption{\label{result-xnli} Accuracy scores on XNLI dataset. The results with $\dag$ are taken from  \citet {conneau-2020-unsupervised}.}
        \end{table*}        

\section{Experiments}
\subsection {Experiment Setup}

\begin{figure*}[t!] 
  \centering
  \subfigure{
  \begin{minipage}[c]{0.49\linewidth}
  \includegraphics[width=1\linewidth]{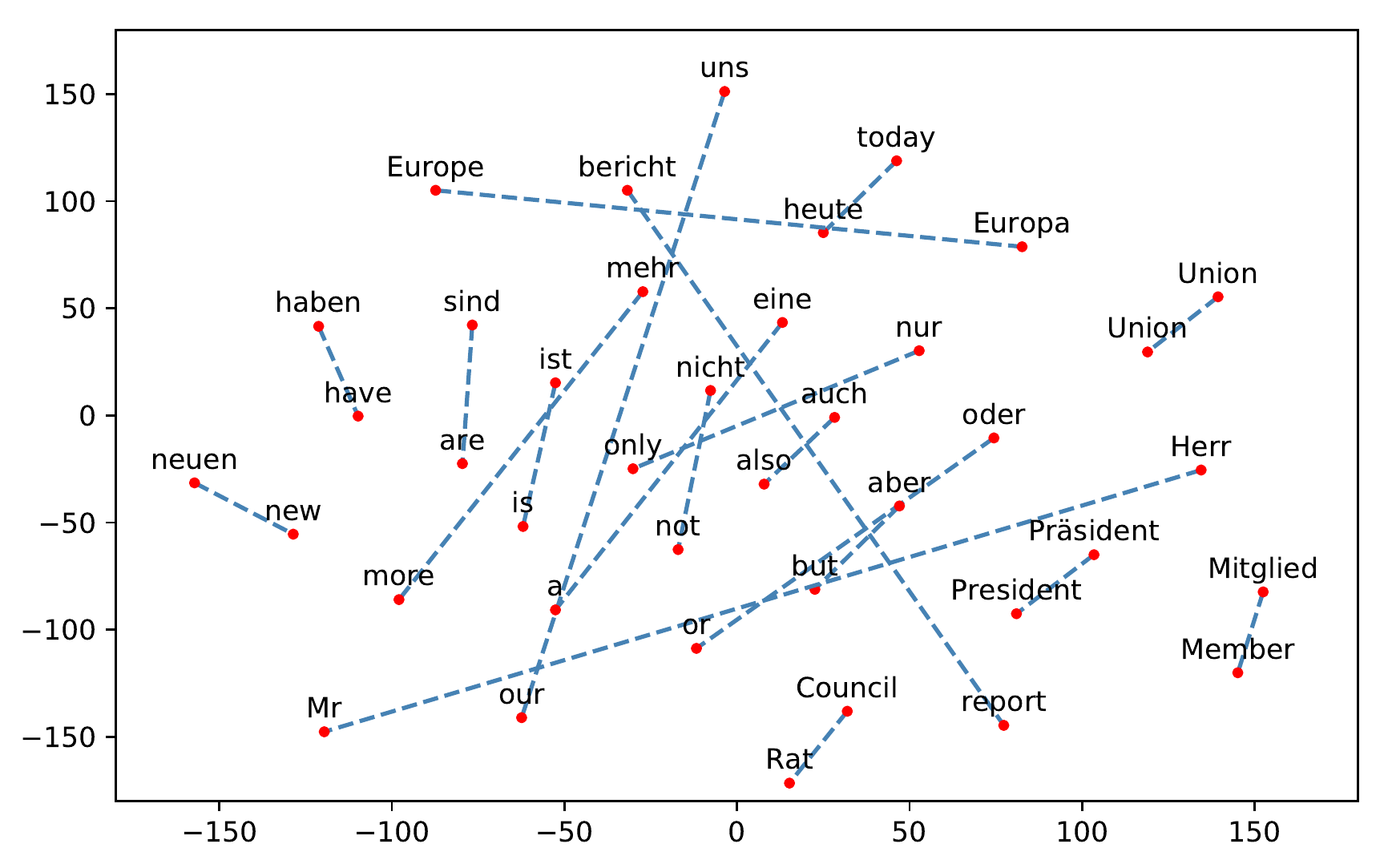}
  \end{minipage}}
  \subfigure{
  \begin{minipage}[c]{0.49 \linewidth}
  \includegraphics[width=1\linewidth]{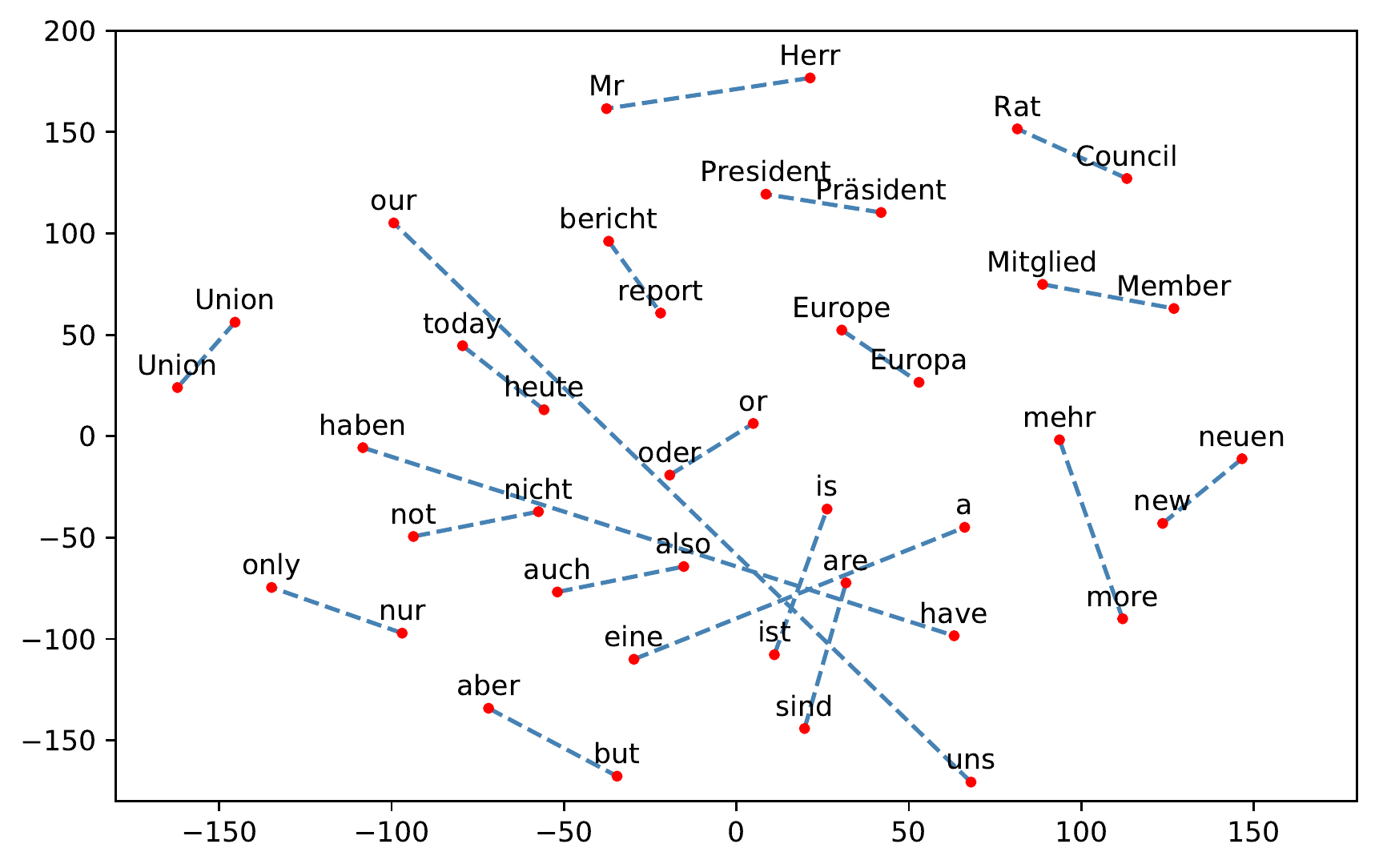}
  \end{minipage}}
  \caption{\label{visual-emb} A visualization of the word embeddings from mBERT before and after WEAM pre-training.  We select 20 English-German word alignment pairs that appear most frequently in the pre-training corpus. Each word alignment pair is connected by a blue dotted line. All the word pairs are identified by FastAlign \citep {dyer-2013-simple}. }
  \end{figure*}
  
We use three parallel corpora with the source language English and the target languages Chinese \footnote{We use the corpus from \citet {bright-xu-2019}.}, German and Spanish \footnote {http://www.statmt.org/europarl} respectively. We initialize the mBERT model with the weights released by Google\footnote{https://github.com/google-research/bert}. We pre-train three models for the three target languages separately to avoid alignment interference among different language pairs.

During the pre-training steps, we empirically set the masking probability as 0.3. Experimentally we find that 0.3 gives better performance. The other settings for masking are the same as the MLM \cite {devlin-2019-bert}.
The hyper-parameters of the three models are the same: we set the learning rate as 5e-5, the batch size as 32, the max sequence length as 128, and the number of pre-training epochs as 2. We set $\lambda$ to 1.

For the downstream evaluation, we fine-tune and test our pre-trained model along with several baselines on the MLQA and XNLI tasks respectively. The specific settings of baselines are described in the following section. 
Since in this work we mainly focus on evaluating the zero-shot performance, we fine-tune all the models in the zero-shot setting where only the English training set is available. We also fine-tune mBERT in the translate-train setting for comparison.

\subsection {Baselines}
We use mBERT \cite{devlin-2019-bert} as our main baseline, which consists of 12 transformer layers, with a hidden size of 768 and 12 attention heads.
For a fair comparison, we also include a baseline mBERT+TLM with the same pre-training settings but uses TLM as the pre-training task.
An additional baseline word-aligned mBERT from \citet {cao-2020-multilingual} is included for the XNLI dataset.

\subsection {Results on MLQA}
Table \ref {result-mlqa} shows our results on MLQA. Note that the results on the target languages of the TLM and WEAM are from models of different language pairs as introduced in the experiment setup section. The results of TLM and WEAM on English are the average of the three models.

The mBERT+TLM model outperforms mBERT by a large margin in the zero-shot setting, but is not as good as the mBERT in the translate-train setting. Our model mBERT+WEAM improves the scores in the zero-shot setting and also outperforms mBERT in the translate-train setting. This result is promising, as it indicates that a properly aligned pre-training model can exceed the performance of translate-train even with zero-shot training.
        
\subsection {Results on XNLI}
Table \ref {result-xnli} shows our results on XNLI.
The mBERT+TLM and word-aligned mBERT achieved similar improvements on this task compared to mBERT, whereas mBERT+WEAM has significantly outperformed both of them. Because all of these models are pre-trained with the same parallel corpus, the differences in performance indicate the importance of considering both the word-level and contextual-level alignment.
Compared with the translate-train result, the mBERT+WEAM result is slightly lower but is close. This is different from MLQA. This observation may indicate that the examples in XNLI have shorter input sequences and thus have fewer translation noises.

 \section{Visualization}

 The effect of contextual alignment has been well studied in \citet {cao-2020-multilingual}, where the authors demonstrate that the contextual alignment is powerful in improving the transferability of mBERT. but the effect of the word-level information alignment is still unclear.
To further explore this problem, we use t-SNE \cite {maaten-2008-visualizing} to visualize the distances between embeddings of word alignment pairs with the highest frequencies (excluding stop words). The result is illustrated in Figure \ref {visual-emb}.

The left panel shows word pairs in the embedding layer of mBERT without WEAM pre-training, we can see that these word pairs are partly aligned. For example, the pairs \textit{today-heute}, \textit{Council-Rat} are aligned well, but \textit{Beriche-report}, \textit{Mr-Herr} are distant away. 
As a comparison, we show the word pairs from the embedding layer of mBERT with WEAM pre-training in the right panel, where most of the word pairs are aligned much better. There are also words that remained poorly aligned even with WEAM. For example, \textit{our-uns}, which may be due to that they are not the exact translation pair (\textit{us-uns} are more exact pairs in this case).
In general, the embeddings are aligned much better after the WEAM pre-training procedure.

 \section{Conclusion}
 In this paper, we propose a new pre-training task named WEAM to align the contextual representations and static word embeddings from multilingual pre-trained models. WEAM consists of a multilingual prediction task and a cross-lingual prediction task.
As a supplement to previous works MMLM or TLM, WEAM introduces the statistic alignment information as prior knowledge to guide the cross-lingual prediction.
Through the experiments on MLQA and XNLI, we show that WEAM can improve the transfer performance significantly and align the word embeddings within the models much better. 
In the future, we plan to extend our method to other multilingual models like XLM-R.

\bibliographystyle{acl_natbib}
\bibliography{Bilingual_Alignment_MRQA_cr}

\begin{thebibliography}{13}
\expandafter\ifx\csname natexlab\endcsname\relax\def\natexlab#1{#1}\fi

\bibitem[{Cao et~al.(2020)Cao, Kitaev, and Klein}]{cao-2020-multilingual}
Steven Cao, Nikita Kitaev, and Dan Klein. 2020.
\newblock Multilingual alignment of contextual word representations.
\newblock \emph{arXiv preprint arXiv:2002.03518}.

\bibitem[{Chi et~al.(2021)Chi, Dong, Wei, Yang, Singhal, Wang, Song, Mao,
  Huang, and Zhou}]{chi-etal-2021-infoxlm}
Zewen Chi, Li~Dong, Furu Wei, Nan Yang, Saksham Singhal, Wenhui Wang, Xia Song,
  Xian-Ling Mao, Heyan Huang, and Ming Zhou. 2021.
\newblock \href {https://doi.org/10.18653/v1/2021.naacl-main.280} {{I}nfo{XLM}:
  An information-theoretic framework for cross-lingual language model
  pre-training}.
\newblock In \emph{Proceedings of the 2021 Conference of the North American
  Chapter of the Association for Computational Linguistics: Human Language
  Technologies}, pages 3576--3588, Online. Association for Computational
  Linguistics.

\bibitem[{Conneau et~al.(2020)Conneau, Khandelwal, Goyal, Chaudhary, Wenzek,
  Guzm{\'a}n, Grave, Ott, Zettlemoyer, and
  Stoyanov}]{conneau-2020-unsupervised}
Alexis Conneau, Kartikay Khandelwal, Naman Goyal, Vishrav Chaudhary, Guillaume
  Wenzek, Francisco Guzm{\'a}n, Edouard Grave, Myle Ott, Luke Zettlemoyer, and
  Veselin Stoyanov. 2020.
\newblock \href {https://doi.org/10.18653/v1/2020.acl-main.747} {Unsupervised
  cross-lingual representation learning at scale}.
\newblock In \emph{Proceedings of the 58th Annual Meeting of the Association
  for Computational Linguistics}, pages 8440--8451, Online. Association for
  Computational Linguistics.

\bibitem[{Conneau and Lample(2019)}]{conneau-2019-cross}
Alexis Conneau and Guillaume Lample. 2019.
\newblock Cross-lingual language model pretraining.
\newblock In \emph{Advances in Neural Information Processing Systems}, pages
  7059--7069.

\bibitem[{Conneau et~al.(2018)Conneau, Rinott, Lample, Williams, Bowman,
  Schwenk, and Stoyanov}]{conneau-2018-xnli}
Alexis Conneau, Ruty Rinott, Guillaume Lample, Adina Williams, Samuel Bowman,
  Holger Schwenk, and Veselin Stoyanov. 2018.
\newblock Xnli: Evaluating cross-lingual sentence representations.
\newblock In \emph{Proceedings of the 2018 Conference on Empirical Methods in
  Natural Language Processing}, pages 2475--2485.

\bibitem[{Devlin et~al.(2019)Devlin, Chang, Lee, and
  Toutanova}]{devlin-2019-bert}
Jacob Devlin, Ming-Wei Chang, Kenton Lee, and Kristina Toutanova. 2019.
\newblock Bert: Pre-training of deep bidirectional transformers for language
  understanding.
\newblock In \emph{Proceedings of the 2019 Conference of the North American
  Chapter of the Association for Computational Linguistics: Human Language
  Technologies, Volume 1 (Long and Short Papers)}, pages 4171--4186.

\bibitem[{Dyer et~al.(2013)Dyer, Chahuneau, and Smith}]{dyer-2013-simple}
Chris Dyer, Victor Chahuneau, and Noah~A Smith. 2013.
\newblock A simple, fast, and effective reparameterization of ibm model 2.
\newblock In \emph{Proceedings of the 2013 Conference of the North American
  Chapter of the Association for Computational Linguistics: Human Language
  Technologies}, pages 644--648.

\bibitem[{Hendrycks and Gimpel(2016)}]{hendrycks-2016-gaussian}
Dan Hendrycks and Kevin Gimpel. 2016.
\newblock Gaussian error linear units (gelus).
\newblock \emph{arXiv preprint arXiv:1606.08415}.

\bibitem[{Huang et~al.(2019)Huang, Liang, Duan, Gong, Shou, Jiang, and
  Zhou}]{huang-2019-unicoder}
Haoyang Huang, Yaobo Liang, Nan Duan, Ming Gong, Linjun Shou, Daxin Jiang, and
  Ming Zhou. 2019.
\newblock \href {https://doi.org/10.18653/v1/D19-1252} {{U}nicoder: A universal
  language encoder by pre-training with multiple cross-lingual tasks}.
\newblock In \emph{Proceedings of the 2019 Conference on Empirical Methods in
  Natural Language Processing and the 9th International Joint Conference on
  Natural Language Processing (EMNLP-IJCNLP)}, pages 2485--2494, Hong Kong,
  China. Association for Computational Linguistics.

\bibitem[{Lewis et~al.(2019)Lewis, O{\u{g}}uz, Rinott, Riedel, and
  Schwenk}]{lewis-2019-mlqa}
Patrick Lewis, Barlas O{\u{g}}uz, Ruty Rinott, Sebastian Riedel, and Holger
  Schwenk. 2019.
\newblock Mlqa: Evaluating cross-lingual extractive question answering.
\newblock \emph{arXiv preprint arXiv:1910.07475}.

\bibitem[{Maaten and Hinton(2008)}]{maaten-2008-visualizing}
Laurens van~der Maaten and Geoffrey Hinton. 2008.
\newblock Visualizing data using t-sne.
\newblock \emph{Journal of machine learning research}, 9(Nov):2579--2605.

\bibitem[{Ouyang et~al.(2020)Ouyang, Wang, Pang, Sun, Tian, Wu, and
  Wang}]{DBLP:journals/corr/abs-2012-15674}
Xuan Ouyang, Shuohuan Wang, Chao Pang, Yu~Sun, Hao Tian, Hua Wu, and Haifeng
  Wang. 2020.
\newblock \href {http://arxiv.org/abs/2012.15674} {{ERNIE-M:} enhanced
  multilingual representation by aligning cross-lingual semantics with
  monolingual corpora}.
\newblock \emph{CoRR}, abs/2012.15674.

\bibitem[{Xu(2019)}]{bright-xu-2019}
Bright Xu. 2019.
\newblock \href {https://doi.org/10.5281/zenodo.3402023} {Nlp chinese corpus:
  Large scale chinese corpus for nlp}.

\end{thebibliography}

\end{document}